\documentclass[11pt,a4paper]{article}
\usepackage[hyperref]{acl2021}
\usepackage{times}
\usepackage{latexsym}

\usepackage{linguex}

\newcommand{\sds}[0]{\textsc{sds}}
\newcommand{\asr}[0]{\textsc{asr}}
\newcommand{\hri}[0]{\textsc{hri}}
\newcommand{\nlu}[0]{\textsc{nlu}}
\newcommand{\nlg}[0]{\textsc{nlg}}

\usepackage{microtype}

\usepackage[pdftex]{graphicx}
\usepackage{enumitem}

\aclfinalcopy 

\title{The State of SLIVAR: What's next for robots, human-robot interaction, and (spoken) dialogue systems?}

\author{Casey Kennington \\
  Department of Computer Science \\
  Boise State University \\
  \texttt{caseykennington@boisestate.edu}
}

\date{}

\begin{document}
\maketitle

\begin{abstract}
We synthesize the reported results and recommendations of recent workshops and seminars that convened to discuss open questions within the important intersection of robotics, human-robot interaction, and spoken dialogue systems research. The goal of this growing area of research interest is to enable people to more effectively and naturally communicate with robots. To carry forward opportunities networking and discussion towards concrete, potentially fundable projects, we encourage interested parties to consider participating in future virtual and in-person discussions and workshops. 
\end{abstract}

\section{Introduction}
Advances in robotics and human-language technologies are accelerating, from robots that vacuum floors to more natural dialogue with Alexa devices. A newly forming \emph{Special Interest Group for Spoken Language Interaction with Virtual Agents and Robots} (SLIVAR) aims to bring these two broad areas together with the goal of empowering people to communicate with robots the way that humans largely communicate with each other: natural human language, particularly spoken dialogue. As robots permeate the world from industry to households, their widespread adoption will be tied to the ability for non-technical, lay people to interact with them as easily as possible. Imagine a robot that can perform its task of vacuuming the floor, but also informing a nearby human that it bumped into an object that happens to be a long-lost TV remote, or that its dust bin is full and needs help being emptied. Better still, robots that can \emph{learn} about their environment and the people with whom they will interact will be more adaptable to many settings and a diversity of users, and what better way to learn about the world than to ask a person?

\paragraph{Definitions} A \emph{robot} is an actuated mechanism programmable in two or more axes with a degree of autonomy, moving within its environment, to perform intended tasks that serve the needs of a user \cite{Eskenazi2020}.\footnote{This, as well as other arguments and definitions that follow are unceremoniously lifted from Roger Moore's excellent keynote at RO-MAN 2020  (\url{https://tinyurl.com/rmoreroman})}$^,$\footnote{See also this Twitter thread: \url{https://twitter.com/BLeichtmann/status/1314080122169970688}}  A \emph{spoken dialogue system} (\sds) is a an automated system that is able to converse with a human with voice.\footnote{Here, the focus is on robots and speech, though in many cases the arguments made here also extend to virtual agents and communication using a text medium.}

\paragraph{Robot complexities} Robots are embodied and can interact with the physical world. Some have wheels, others feet; arms, others lifts; some are designed to appear to have human-like physical characteristics, others are purposefully designed not to. The variance in robot morphology, functionality, and target audience varies widely. Most robots are designed with specific hardware for narrow tasks. Common application areas for robots include industrial robotics, health care robots, and household consumer robots. Design and testing of robots is engineering and time-intensive; robots have constraints on battery power, available sensors, functionality, locomotion, and ease of use. 

\paragraph{Spoken language complexities} Most natural language processing research is text-focused. Language model advancements based on transformers are impressive, but the field often underestimates the richness and complexity of spoken language, especially in a co-located setting with other physical entities such as a robot. As natural as it is for people to engage in dialogue with each other, engaging in dialogue with a machine is not a natural expectation. Spoken interaction is complex and messy. People speak with accents that make speech recognition difficult, not to mention common spoken dialogue artifacts such as false starts, repetitions, prosody, filled time-buying strategies, laughter and other extra-linguistic vocalizations.

To further illustrate how challenging spoken dialogue can be, \newcite{Schlangen2020} gives an example of a fairly mundane dialogue between two people walking together in a park: 

\label{ex:example_description}  
\a. \textbf{Ann}: Look at the dog, over at the water fountain.
\b. \textbf{Bert}: I know, isn’t it cute!
\c. (later, they meet a mutual friend)
\d. \textbf{Bert}: We just saw man carrying a dog! 
\e. \textbf{Ann}: The cutest poodle ever!
\f. \textbf{Bert}: Actually, I don’t think that was a poodle. It was too tall.  I think it was a labradoodle. 

If one of the participants in this dialogue, say Bert, were to be replaced by an automated system, then Bert would need to be able to handle deictic references to external objects (a), identify what a dog is (a), know that poodles and labradoodles are types of dogs to the degree that Bert knows the difference between two similar types and can express confidence in that knowledge (f; requiring negotiation skills, as noted by Schlangen), be able to talk about events that took place in the past about an entity that is no longer perceptually available (d-e), and use anaphoric references such as the word \emph{it} (b). While it is true that developing working systems requires ample data and engineering, handling these types of phenomena together in a single system is \emph{not} low-hanging fruit for research. It is our observation that these kinds of dialogue artifacts are not well-known in the broader research community, to say nothing of higher-level pragmatic issues such as allusions.

\paragraph{Challenges to endowing robots with spoken language abilities} Due to the above-mentioned complexities of robots and spoken language, it is not enough to plug existing \sds\ systems into existing robot platforms and expect any degree of natural, robust, or even useful communication. Focusing just on automatic speech recognition (\asr) as a part of \sds, \asr\ in a setting with a robot is wrought with its own challenges: the robot actuators could make noise and the changing acoustics of the room due to the robot's location could affect speech recognition, to say nothing of the full-duplex nature of co-located spoken interaction; i.e., the robot needs to always be listening even if it's talking or carrying out an action. 

Beyond speech recognition, natural language understanding (\nlu) is challenging even if speech recognition is perfect because people don't usually speak in complete, grammatical sentences and the fact that words and phrases refer to real, physical things that could be co-present with the human and the robot or abstractly refer to a known entity. The \nlu\ plays an important role of mapping from speech or transcription to a computable semantic abstraction, one that is often tied to the task. 

What should the robot say, and when? Should the robot stick to the task or offer freer chat cababilities? Natural language generation (\nlg) is very challenging because what the system says is how the system informs the user of its state or move the dialogue forward towards a shared task goal. It's not enough to say something grammatical, it also needs to be coherent and contribut to the unfolding dialogue. 

\paragraph{Challenges for speech and human-robot interaction} On the same argument that one cannot simply attach the best-performing \asr\ to a robot and expect natural communication, one cannot simply put humans and robots in the same room to work together on some task and expect successful outcomes, even on narrowly-defined tasks. Part of the problem is that the abilities of robots is limited, but the other and perhaps more vexing problem is on the side of the humans: when humans are confronted with robots and told to work with them, even in lab settings, humans immediately have expectations of what the robots should do, and humans also go so far as to anthropomorphize robots for gender, age, social status, among other things (in some cases, only based on how the robots appear, not how they behave), and these expectations and perceptions have implications for the kinds of things robots can do with humans. This intricate area of how humans perceive robots and the nuances of how robots and humans interact with each other (using speech or any other means) is the focus of the field of human-robot interaction (\hri). 

Both \sds\ and robots take actions, and in order to take actions they must make decisions. For a \sds, that is often handled in a dialogue manager or dialogue state tracker. Robots likewise have modules that make high-level decisions on goals and low-level decisions on how to reach those goals, like move an arm or move a camera. Should decisions about what the system should say and what it should do be made jointly, or separately? 

It is our goal that the lessons learned from roboticists, \sds\ researchers (including speech, but also textual modalities) and \hri\ researchers can join forces to work towards humans and robots collaborating on terms that are (at least to some degree) safe and natural for the humans involved. The purpose of this paper is to synthesize what has happened recently in this cross-cutting research area to identify and overcome the complexities, challenges, and open questions to language communication with robots. In the following section, we outline some recent work including workshops and publications that contribute to the discussion, then propose next steps for the growing SLIVAR community. 

\section{Recent Efforts}

Two U.S. National Science Foundation workshops were convened in October, 2019: \emph{Future Directions Workshop, Toward User-Oriented Agents: Research directions and Challenges} \cite{Eskenazi2020} which focused on the role of intelligent agents and how to make them more user-oriented. The participants of the workshop identified broad areas and themes for future directions (taking note of common pitfalls \cite{balentine2007s}) including applications, infrastructure, dynamic views of user-agent interaction, and made several recommendations in building low-cost dialogue systems, multimodal, grounded, and situated interaction, robust and flexible dialogue management, and intelligent agents as good actors. The second workshop \emph{Spoken Language Interaction with Robots} \cite{Marge2020,MARGE2022101255} focused on speech and the complexities thereof when robots are involved. The participants identified recommendations relating to eight different themes: 
\begin{enumerate}
    \item First, meeting human needs requires work on new challenges in speech technology and user experience design. 
    \item Second, this requires better models of the social and interactive aspects of language use.
    \item Third, for robustness, robots need higher-bandwidth communication with users and better handling of uncertainty, including simultaneous consideration of multiple hypotheses and goals. 
    \item Fourth, more powerful adaptation methods are needed, to enable robots to communicate in new environments, for new tasks, and with diverse user populations, without extensive re-engineering or the collection of massive training data.
    \item Fifth, since robots are embodied, speech should function together with other communications  modalities,  such  as  gaze,  gesture,  posture  and  motion.
    \item  Sixth,  since robots operate in complex environments, speech components need access to rich yet efficient representations of what the robot knows about objects, locations, noise sources,the user, and other humans.
    \item  Seventh, since robots operate in real time, their speech and language processing components must also.
    \item Eighth, in addition to more research, we need more work on infrastructure and resources, including shareable software modules and internal interfaces, inexpensive hardware, baseline systems, and diverse corpora.
\end{enumerate}         

A Dagstuhl (Germany) Seminar convened in January 2020 on the topic of SLIVAR.\footnote{Dagstuhl ID 20021 \url{https://drops.dagstuhl.de/opus/volltexte/2020/12400/}} This resulted in organization of other events such as a special session on robots and dialogue (RoboDial 2.0) at the SIGDIAL 2020 conference,\footnote{\url{https://robodial.github.io}} a workshop on natural language generation at the HRI 2020 conference,\footnote{\url{https://hbuschme.github.io/nlg-hri-workshop-2020/}} and a workshop ROBOTDIAL at the IJCAI 2020 conference.\footnote{\url{http://sap.ist.i.kyoto-u.ac.jp/ijcai2020/robotdial/}} The primary goal of the Dagstuhl Seminar was to provide discussion and establish a community. Discussions revolved around ethics, usability, scenarios for human-agent / human-robot groups, evaluation, architectures, and situated language understanding. Other recent, related events include a AAAI Symposium on Natural Communication for Human-Robot Collaboration (2017),\footnote{\url{https://www.ttic.edu/nchrc/}} and RoboNLP workshops including spatial language understanding.\footnote{See and links to prior workshops \url{https://splu-robonlp.github.io/}}

While many have contributed in different ways to the convergence between robotics, \sds, and \hri\ research, two recent articles highlight recent trends, challenges, and opportunities at these important crossroads. \newcite{Tellex2020} surveys the use of natural language from a robotics point of view including natural language issues that are common on robots and current state of the art with focus on tasks such as robot navigation. \newcite{Kragic2018} makes the case that though there has been progress, things like full autonomy, learning in collaboration with people, and safe and flexible performance in varied environments remain elusive. 

\section{Basic Requirements}

\newcite{Kennington2020} identified 5 basic requirements for \emph{robot-ready} \sds. These requirements are largely at the level of framework and processing; not all tasks and applications will need all of the requirements, nor is the list comprehensive (see Opportunities Section below), but a goal that robust and natural dialogue will need and as a starting point for the opportunities that follow in the next section:\footnote{These requirements were derived from discussions in the above-mentioned workshops.}

\begin{enumerate}[noitemsep, label=\textbf{\arabic*}.]
    \item \textbf{modular}: robot components are modular and individual modules must be able to integrate with \sds\ modules 
    \item \textbf{multimodal}: robots are \emph{situated} dialogue partners having many sensors along with the \sds\ speech input; those sensors must be integrated together
    \item \textbf{distributive}: robot and \sds\ modules are often computationally expensive; modules should be able to easily communicate with each other in a distributed environment
    \item \textbf{incremental}: modules must be able to process input quickly and immediately 
    \item \textbf{aligned}: sensors must be temporally aligned; i.e., synchronized in time
\end{enumerate}

\section{Opportunities}

\subsection{Infrastructure}

\begin{itemize}[noitemsep]
    \item Common Tools and Platforms: A common research platform should be developed or extended for
intelligent agent research. 
    \item Corpora: The collection of situated and multimodal corpora is an important future direction.
    \item Shared Tasks: With a shared infrastructure and corpora, shared tasks will 
promote effective and comparable research. 
    \item Evaluation: Metrics need to be designed that account for the nuances of dialogue with intelligence agents like robots.
    \item Science and Engineering: Work to elucidate the fundamental questions in real-time social interaction, both scientific and engineering.
\end{itemize}

\subsection{Multimodal, grounded and situated interaction}

\begin{itemize}[noitemsep]
    \item The range of modalities used in today's dialog systems should be broadened, for example by including prosody, haptics, or motion capture as input modalities. 
    \item Situated dialog should play a central role in future dialog systems research. 
    \item The advances of continuous (i.e., incremental) speech processing should be generalized to incremental multimodality. 
\end{itemize}

\subsection{Audio and Speech Processing}

\begin{itemize}[noitemsep]
    \item Better exploit context and expectations in speech recognition.
    \item Consider creating a speech recognition system focused on the issues encountered in robotics (e.g., splitting speech signals and speech in noise where the noise may come from the robot and/or the surrounding environment).
    \item Better exploit prosody, emotion, and mental state from the speech signal.
    \item Use audio scene and sound event analysis to better understand the environment.
\end{itemize}

\subsection{Mutual understanding between robot and human}

\begin{itemize}[noitemsep]
    \item Develop language understanding models for robots that resolve referential ambiguities, particularly to physically-present objects and locations in situated dialogue. 
    \item Dynamically increase the fidelity of how robots represent the state of the human interlocutor.
    \item Inference and understanding should be based on the user’s goals and beliefs.
    \item Explore the broader space of clarification and recovery strategies in spoken language interaction with co-present agents, including when and how. 
    \item Research should be highly interdisciplinary, including linguistics, pragmatics and reasoning.
    \item Focus on language not only as a way to achieve human-like behaviors, but also as a way to support limited but highly usable communications abilities.
    \item Work to better characterize the list of communicative competencies most needed for robots in various scenarios
    
\end{itemize}

\begin{figure*}[t]
    \centering
    \includegraphics[width=.8\textwidth]{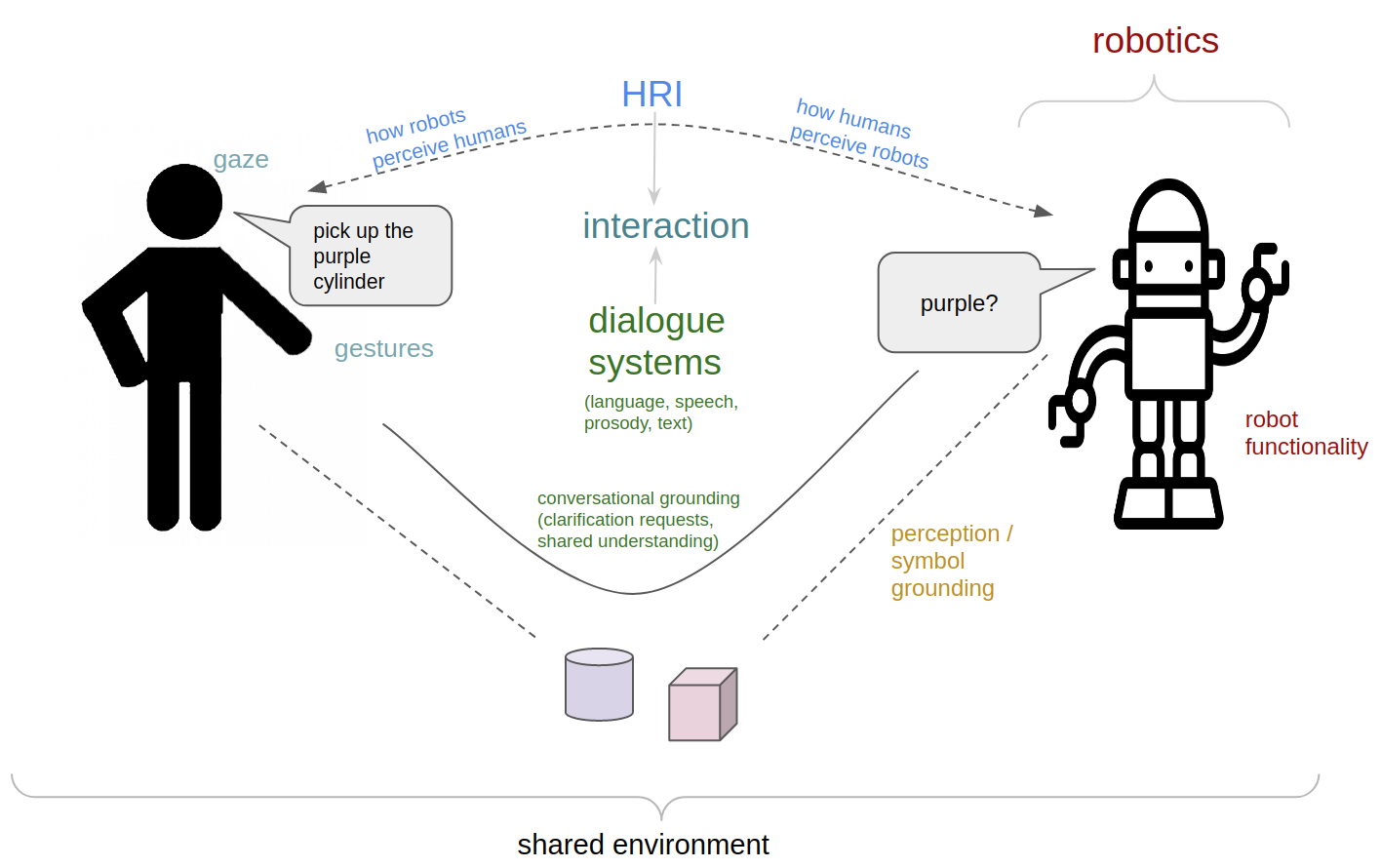}
    \caption{The fields of robotics, \hri, and \sds\ have different, yet complementary goals.}
    \label{fig:candidate_generation_success}
\end{figure*}

\subsection{Robust and Flexible}

\begin{itemize}[noitemsep]

    \item Research should be inspired by human dialog techniques in order to efficiently address others
goals. Discourse context (as well as physical context) should be preferred to making explicit representations.
    \item Explicit representations of context should be designed in a way that enables them to be shared
with other domains, tasks and applications.
    \item Agents should support different user style of interaction preferences
    \item Include partially-redundant functionality.
    \item Make components robust to uncertainty.
    \item Focus  not  only  on  improving  better  core  components,  but  also  on  cross-cutting issues and those that have fallen between the cracks
    \item Make systems and components adaptable to users.
\end{itemize}

\subsection{Applications}

\begin{itemize}[noitemsep]
    \item Applications are often narrow, but should be as generalizable as possible.
    \item Applications are dependant on the target user pool / audience. 
    \item Applications should be sensitive to aspects of the individual user (e.g. their satisfaction, their cognitive or behavioral.
change, their speaking proficiency)
\end{itemize}

\subsection{Dynamic Views of User-Agent Interaction}

\begin{itemize}[noitemsep]
    \item Move research from generic, static views of the user, the agent and their relationship to personalized and dynamic models
    \item Move from human-computer to multi-party, where the agent learns about different humans.
    \item Research emphasize scenarios that involve longer conversations (beyond 1-2 turns) and/or more frequent interactions.
from the same user or groups of users.
    \item Deliberately engineer user perceptions and expectations.
\end{itemize}

\subsection{Building Dialogue Systems in Low-Resource Conditions}

\begin{itemize}[noitemsep]
    \item A systematic way to introduce prior knowledge as an input to data-driven models.
    \item Community effort to make pretraining and transferable models well-documented, thoroughly tested and widely available.
    \item Open-source, robust and reusable tools for solved tasks, e.g. wizard-of-oz interface, data
collection tools, human evaluation platform).
    \item A principled approach for system design, data collection, resources management, and evaluation for new AI projects with limited/no data
    \item New machine learning methods that can better utilize knowledge from other domains or unannotated data.
    \item Shared tasks that standardize the evaluation of dialog agents under low resources setting, and
regular benchmark comparison across various methods.
\end{itemize}

\subsection{Intelligent Agents as Good Actors}

\begin{itemize}[noitemsep]
    \item Research plans within scientific proposals should raise and address at least one ethical issue.
    \item Education of students and newcomers to the field on ethical issues. 
    \item Develop outreach efforts for increasing public awareness in collaboration with target user demographics.
    \item For every NSF/government proposal or grant that includes the release of data, the data should
have a data sheet.
\end{itemize}

\subsection{Policy}

\begin{itemize}[noitemsep]
    \item Fund spoken language interaction with robots as its own area of research. 
    \item Prefer evaluation-based on use cases.
    \item Support many kinds of research and development activities.
    \item Work to overcome the barriers to data sharing.
    \item Explore novel public-private partnerships for open source software
\end{itemize}

\section{Discussion}

Taken together, we highlight some of the opportunities and complementary goals from within the fields of \textsc{sds}, \textsc{hri}, and robotics in Figure~\ref{fig:candidate_generation_success}. The figure shows a human interacting with a robot in a co-located environment with two objects; the dashed lines show perceptions, solid lines and arrows show relations; colors denote areas of research that has been conducted in the respective fields which complement the goals of SLIVAR. 

There are ample opportunities to collaborate and work on gratifying and impactful projects at the human-robot frontier. We intend to convene a workshop in the near future to work towards these goals by building relationships and identifying specific research projects. 

\paragraph{Acknowledgements} I want to thank Maxine Eskenazi for helpful feedback and suggestions.

\bibliographystyle{acl_natbib}
\bibliography{refs}

\end{document}